\documentclass[10pt,twocolumn,letterpaper]{article}

\usepackage[pagenumbers]{cvpr} % To force page numbers, e.g. for an arXiv version
\usepackage{multirow}
\usepackage{array}
\definecolor{cvprblue}{rgb}{0.21,0.49,0.74}
\usepackage[pagebackref,breaklinks,colorlinks,allcolors=cvprblue]{hyperref}

\def\paperID{*****} % *** Enter the Paper ID here
\def\confName{CVPR}
\def\confYear{2026}

\title{From Pixels to Semantics: A Multi-Stage AI Framework for Structural Damage Detection in Satellite Imagery}

\author{Bijay Shakya$^*$, Catherine Hoier$^*$, Khandaker Mamun Ahmed\textsuperscript{\textdagger}\\
The Beacom College of Computer \& Cyber Sciences, Dakota State University, USA \\
{\tt\small \{Bijay.Shakya@trojans, Catherine.Hoier@trojans, khandakermamun.Ahmed@\}.dsu.edu}
}

\begin{document}
\maketitle
\let\thefootnote\relax\footnote{$^*$ Equal Contributions.}\\
\let\thefootnote\relax\footnote{\textsuperscript{\textdagger}Corresponding author.}

\begin{abstract}
Rapid and accurate structural damage assessment following natural disasters is critical for effective emergency response and recovery. However, remote sensing imagery often suffers from low spatial resolution, contextual ambiguity, and limited semantic interpretability, reducing the reliability of traditional detection pipelines. In this work, we propose a novel hybrid framework that integrates AI-based super-resolution, deep learning object detection, and Vision-Language Models (VLMs) for comprehensive post-disaster building damage assessment. First, we enhance pre- and post-disaster satellite imagery using a Video Restoration Transformer (VRT) to upscale images from $1024$×$1024$ to $4096$×$4096$ resolution, improving structural detail visibility. Next, a YOLOv11-based detector localizes buildings in pre-disaster imagery, and cropped building regions are analyzed using VLMs to semantically assess structural damage across four severity levels. To ensure robust evaluation in the absence of ground-truth captions, we employ CLIPScore for reference-free semantic alignment and introduce a multi-model VLM-as-a-Jury strategy to reduce individual model bias in safety-critical decision-making. Experiments on subsets of the xBD dataset, including the Moore Tornado and Hurricane Matthew events, demonstrate that the proposed framework enhances the semantic interpretation of damaged buildings. In addition, our framework provides helpful recommendations to first responders for recovery based on damage analysis.
\end{abstract}    
% \ahmed{venue: https://sites.google.com/view/ai4rwc2026/home?authuser=0}
% \ahmed{Submission deadline: March 15, 2026}
% \ahmed{https://eccv.ecva.net/Conferences/2026/Dates, Submission Deadline	Mar 05 '26}

%\input{sec/1_intro}
\section{Introduction}
\label{sec:intro}

\begin{figure}
    \centering
    \includegraphics[width=1.0\linewidth]{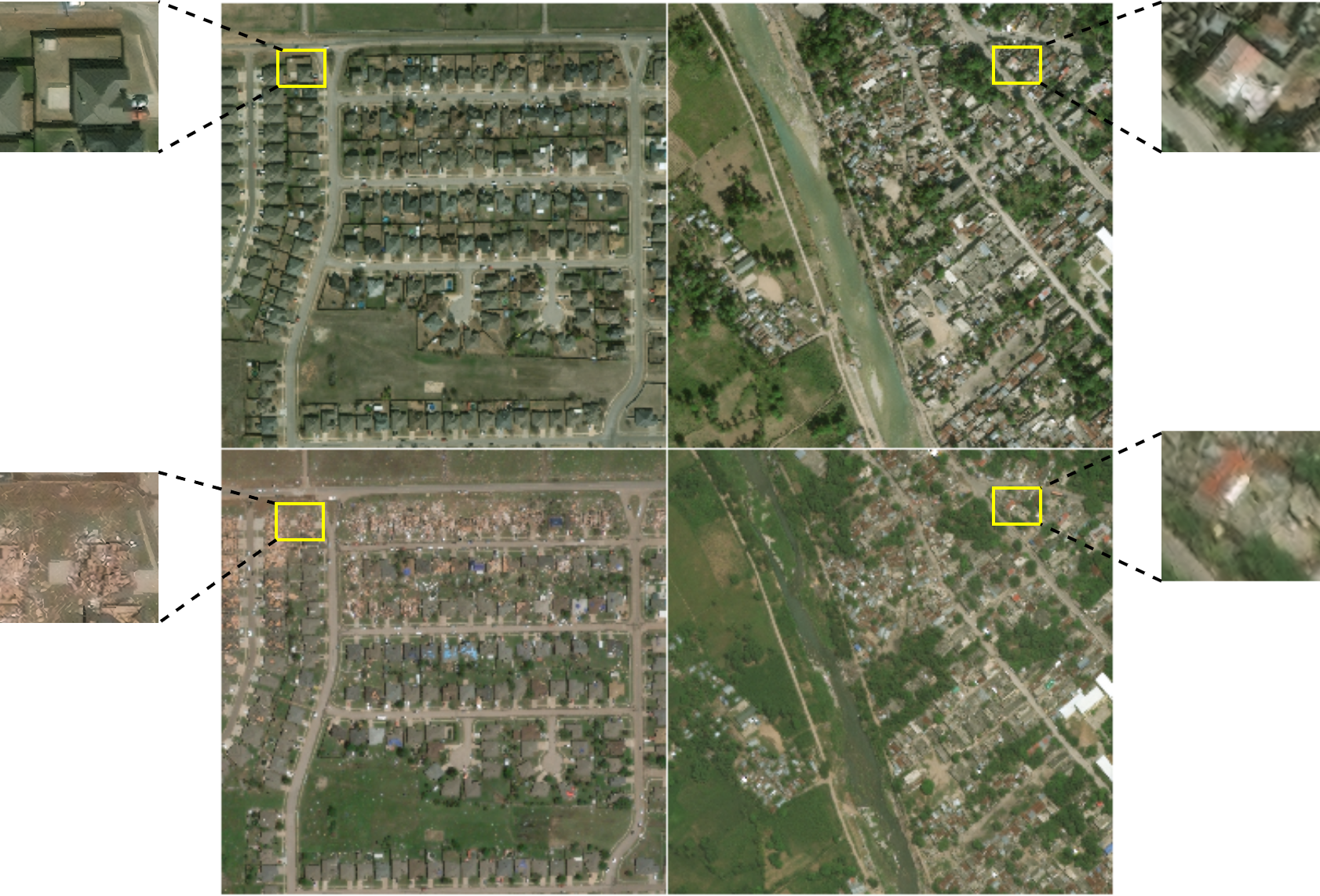}
    \caption{xBD dataset samples: pre-disaster images (top) and post-disaster images (bottom). From left to right: Tornado Moore; Hurricane Matthew.}
    % \ahmed{Please update this figure. It is not showing a significant difference between the pre and post disaster}
    % \bijay{I checked other images in the xBD dataset. Most of them do not show significant changes due to the high altitude angle. Even the images in the xBD paper are not clear. Ref: https://arxiv.org/pdf/1911.09296}
    \label{fig:dataset_samples}
\end{figure}

Natural disasters are unpredictable events that often occur without warning, causing significant risk to human life and property. Various disasters such as earthquakes, tornadoes, hurricanes, and floods have often been reported, leading to considerable loss of life, severe destruction of infrastructure, and major economic disruptions in various parts of the world. According to the UNDRR Global Assessment Report (GAR) 2025, earthquakes account for more than a quarter (25.6\%) of global economic disaster losses \cite{undrr2025gar}. Recent earthquake events, such as a powerful 7.1 magnitude earthquake, struck at approximately 620km depth, centered about 55km NNW of Kota Belud, Malaysia \cite{quakepulse_us6000sasz_2026}.
The 2023 Kahramanmaraş earthquake sequence resulted in the deaths of over 50,000 individuals in Türkiye; the Myanmar earthquake in March-April 2025 reported over 3,700 fatalities and over 5,100 injuries, with extensive damage to infrastructure \cite{undrr2025gar, unhcr_myanmar_earthquake_onemonth_2025}. On February 19, 2026, Central Indiana was hit by severe storms, resulting in multiple tornadoes and extensive damage, particularly south of Interstate 70. The National Weather Service confirmed two tornadoes, classified as EF2 and EF0, which inflicted widespread destruction in the area~\cite{nwsind_feb19_2026_severe}. Likewise, 1,796 tornadoes have been preliminarily confirmed across the U.S.A in 2024, marking the second-highest total record after 2004 \cite{noaa2024tornado}. Approximately 90 deaths have been confirmed globally due to various tornado events, with the majority of fatalities reported in the United States, accounting for 52 fatalities in 2024 \cite{statista_tornado_fatalities_us_1995_2023}. Furthermore, intense tropical cyclone Gezani struck eastern Madagascar on February 10, 2026, with winds of about 180 km/h. The cyclone caused severe damage in Toamasina, affecting nearly $270,000$ individuals, displacing around 16,000 people, and resulting in at least 40 fatalities \cite{wmo_gezani_madagascar_mozambique_2026}. These staggering figures highlight the importance of effective disaster assessment and mitigation strategies, ensuring quick recovery. However, traditional and manual assessment methods are often labor-intensive and time-consuming, which causes delays in recovery efforts and stretches limited resources. In addition, even if access to affected locations may be possible, dangerous conditions such as extreme heat and unstable structures pose serious safety risks to rescuers \cite{perce2007disaster}.

To address these limitations, researchers are using advanced technologies to gather information on buildings impacted by natural disasters. Various state-of-the-art techniques for data collection, remote sensing, and image-based analysis for precise post-disaster building assessment \cite{al2024integrating}. The research \cite{jiang2019building, malmgren2020sentinel} utilized remote sensing techniques for detecting building and infrastructure damage using Space-borne synthetic aperture radar (SAR) images. The studies \cite{martinelli2023damage,li2019building} leveraged object detection methods like You Only Look Once (YOLO), Single Shot Multibox Detector (SSD) for accurate building damage assessment on aerial imagery. Likewise, Visual Language Model-based methods are proposed to aid the responders for post-disaster damage assessment through visual question answering in the research work \cite{sarkar2021vqa, karimi2025zeshot}. However, these methods pose their own bottlenecks, like remote sensing methods struggle with data quality and label noise issues; object detection methods lack contextual understanding and can only detect what they are trained for, and VLM-based methods often experience hallucination problems in high-resolution images. Therefore, it necessitates a hybrid approach to generate a precise assessment in safety-critical contexts. To overcome these bottlenecks, our research work presents a hybrid disaster assessment framework that leverages super-resolution for enhanced data quality, employs YOLO-based building damage detection, and incorporates visual reasoning for accurate post disaster damage evaluation.

The main contributions of our work are as follows:

\begin{itemize}[nolistsep, leftmargin=1.5em]
    \item \textbf{Super-resolution Enhancement:} We employ a super-resolution model to enhance pre-and post-disaster video footage quality, thereby improving damage detection accuracy.
    \item \textbf{Deep Learning-based Damage Detection:} We train and utilize the YOLOv11 model to detect affected buildings in the provided footage. 
    \item \textbf{VLM-based Damage Assessment:} We leverage VLM responses to assess building structural damage and evaluate their reliability through standard metrics: CLIPScore, F1-Score, and aggregated VLM judge scores, aiding the rescue team for recovery efforts.
\end{itemize}

% The remainder of this paper is organized as follows. Section \ref{sec:related_works} reviews related work in structural damage detection and post-disaster assessment. Section 3 details the proposed framework for post-disaster assessment, covering data collection, video enhancement, model-specific damage detection, and VLM-based analysis. Section 4 discusses the dataset used in this study. Section 5 presents the experimental results and key findings. Finally, Section 6 outlines the concluding remarks and highlights possible future research directions.

\section{Related Works}
\label{sec:related_works}

\textbf{Super-Resolution for Improved Damage Detection.}
Data quality is critical for reliable structural damage detection. Low-resolution imagery often fails to capture fine-grained details such as cracks, deformations, and partial collapses, leading to inaccurate damage assessment \cite{al2024integrating}. Post-disaster assessment frequently relies on aerial and satellite imagery, which typically contains background noise, multi-scale objects, and illumination variations that can further degrade detection performance if not properly processed. Super-resolution (SR) techniques have therefore been widely adopted to enhance spatial details and improve downstream damage detection. Prior studies demonstrate that SR improves the detection of fine structural defects in concrete and foundations \cite{kim2023learning, pawar2025deep}, pavement damage \cite{pham2024improving, inzerillo2022super, yuan2022super}, and failures in critical infrastructure such as power lines \cite{zhou2024srgan}.

Several works integrate SR directly into damage assessment pipelines. For example, \cite{fu2022toward} address the limited availability of high-resolution post-disaster imagery caused by long satellite revisit times by employing a Super-Resolution Generative Adversarial Network (SRGAN) to reconstruct high-resolution images from low-resolution inputs. The reconstructed images are then processed by a U-Net-based damage detection model, improving building boundary delineation and detection accuracy. Similarly, \cite{kim2023learning} demonstrate that SRGAN-enhanced imagery significantly improves crack detection in blurred or low-resolution images. To address class imbalance in post-disaster datasets, \cite{lagap2025enhancing} employ Enhanced-SRGAN as a preprocessing module to improve the spatial quality of minority-class samples, leading to more reliable damage detection from low-resolution satellite imagery. 

\textbf{Deep Learning for Structural Damage Detection.} In recent years, deep learning and computer vision have significantly advanced structural health monitoring by enabling rapid, scalable, and accurate damage assessment \cite{ai2023computer}. Because structural damage evaluation is a safety-critical task, manual inspection or direct access to affected sites can be hazardous, particularly in post-disaster environments. Deep-learning-based approaches mitigate these risks by automatically analyzing images and video footage of damaged infrastructure. Numerous studies have demonstrated the effectiveness of these methods for detecting damage in buildings, bridges, and roadways. For instance, \cite{raushan2025damage} proposed an automated framework for detecting damage in concrete structures using the YOLO family of models (v3–v10), showing that YOLOv8 and YOLOv10 achieve an effective balance between computational efficiency and detection accuracy, making them suitable for real-time monitoring. Similarly, \cite{pratibha2023deep} employed YOLOv5 to detect cracks in masonry structures using bounding box annotations, achieving a mean average precision (mAP${0.5}$) of approximately 92\%. UAV-based damage localization using YOLO was also explored in \cite{martinelli2023damage}, which reported an mAP${0.5}$ of 0.924 for detecting structural damage across different building types. Beyond YOLO-based approaches, \cite{li2019building} improved the Single Shot Multibox Detector (SSD) framework for post-disaster building damage detection using the Hurricane Sandy dataset, achieving an mAP of 77.27 when combined with data augmentation techniques. Additionally, \cite{umeike2024accelerating} compared deep learning models for post-tornado damage assessment, showing that YOLOv11 enables real-time detection with 60.83\% accuracy, while a fine-tuned ResNet50 model achieves higher accuracy of 90.28\%. Collectively, these studies demonstrate the effectiveness and practical applicability of deep learning techniques for automated structural damage detection in disaster response and infrastructure monitoring. 

\textbf{VLM for Post-Disaster Damage Assessment.} Visual–Language Models (VLMs) have recently emerged as powerful tools for post-disaster damage assessment due to their ability to jointly reason over visual and textual information. These models can follow task instructions, answer natural-language queries, and generalize across diverse disaster scenarios, making them suitable for emergency response applications. For example, \cite{wang2025disasterm3} introduced \textit{DisasterM3}, a large-scale dataset containing 26,988 bi-temporal satellite images and 123,000 instruction pairs across 36 disaster events, enabling comprehensive evaluation of VLMs for disaster-related perception and reasoning tasks. Similarly, \cite{karimi2025zeshot} proposed a zero-shot VLM-based visual question answering framework for disaster assessment, demonstrating strong adaptability without additional fine-tuning. Similarly, the study \cite{sarkar2021vqa} proposed a visual question answering benchmark for post-disaster damage assessment, providing a comprehensive scene-level understanding.

Despite these advances, disaster imagery collected via UAVs often suffers from low resolution, occlusion, and small object sizes, limiting accurate interpretation. Super-resolution techniques help to enhance the image resolution; however, they alone cannot localize the target objects or provide semantic damage severity. Therefore, object detection models are employed to provide the precise spatial localization information. These localization details, like bounding boxes and object categories, can aid the VLM models in performing high-level semantic reasoning and analyzing damage levels. Motivated by these complementary strengths, we propose a hybrid framework that integrates super-resolution, object detection, and VLM reasoning to improve image quality, localization accuracy, and semantic interpretation for reliable building damage assessment in post-disaster scenarios.
\section{xBD Dataset}
The xBD dataset \cite{gupta2019xbd} is the first large-scale high-resolution satellite imagery dataset, serving as a canonical benchmark for building damage assessment. The dataset includes pre-and post-disaster imagery for six types of disasters with building polygons, damage level labels, and relevant satellite metadata verified by various disaster response agencies. It consists of 850,736 building annotations covering a 45,362 $\text{km}^2$ area across 15 countries. Each image has a Ground Sample Distance (GSD) of 0.5m. 
Besides metadata information, it also provides a standardized four-level qualitative damage scale: No Damage, Minor Damage, Major Damage, and Destroyed for building damage assessment \cite{gupta2019xbd}. 
The dataset primarly focus on the ``Building" category with over 316,000 instances specifically categorized by damage level. Since our research focuses on post-disaster building damage assessment, xBD is the most suitable dataset for our experiment. For damage assessment, we have leveraged the subsets of pre-and post-disaster images for two disasters, namely, the Moore Tornado and Hurricane Matthew. The Moore tornado subset includes 227 sets of pre- and post-images, totaling 454 images. The Hurricane Matthew subset has 238 sets of pre- and post-images, amounting to 476 images. The original image quality for both the tornado and hurricane is 1024 x 1024 pixels, which is upscaled to 4096 x 4096 pixels using the super-resolution method before prediction. Figure  \ref{fig:dataset_samples} illustrates sample pre-and post disaster images of Hurricane Matthew and the Moore Tornado event from the xBD dataset, along with a few cropped buildings.
\section{Methodology}

\begin{figure}
    \centering
    \includegraphics[width=1.0\linewidth]{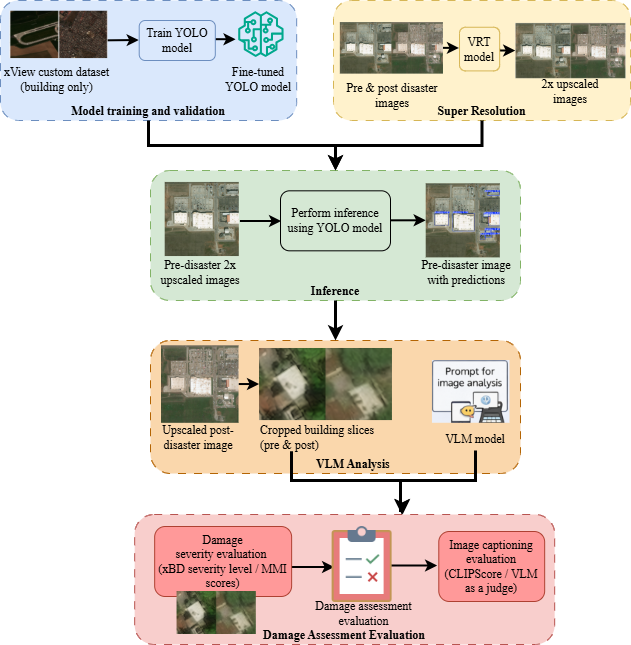}
    \caption{Workflow of the proposed multi-stage structural damage assessment framework. The framework combines AI-driven super-resolution to improve low-resolution satellite images, utilizes a YOLOv11 model for building detection, and employs a Vision-Language Model for assessing damage after disasters through pre- and post-event images.}
    \label{fig:flow_diagram}
\end{figure}

To address the aforementioned challenges in the existing building damage assessment research, we proposed a hybrid building damage assessment framework that integrates various modules to automate the building damage severity classification and provide semantic damage analysis, aiding first responders in recovery tasks.\\\\
\textbf{Data Preparation.} To evaluate our proposed framework, we utilized a satellite imagery subset for two disasters: the Moore Tornado and Hurricane Matthew from the xBD dataset \cite{gupta2019xbddatasetassessingbuilding}. The Moore tornado subset consists of 227 pre- and post-image sets, totaling 454 images, while the Hurricane Matthew subset includes 238 sets, amounting to 476 images. The data preparation involved separating pre- and post-disaster images into separate folders, with each containing two identical copies of the respective images. This organization ensures that the data pipeline is aligned correctly with the super-resolution enhancement algorithm.

For the YOLOv11 model training, we tailored the xView dataset to include only the ``Building" category out of 60 categories, as our research focuses on critical building damage assessment. We chose the xView dataset since we utilize the xBD dataset for our evaluation, which is the updated xView dataset version designed for damage assessment tasks.\\\\
%NOTE: removed the line after the citation: Though typically used in videos for video enhancement, throughout the author's testing of several image enhancement algorithms they have found that using VRT for the satellite imagery provides the best results for YOLOv11 detection and overall VLM analysis.
\textbf{Proposed Method.} Our framework automates the building damage analysis task in three stages. Firstly, we upscale the xBD image subsets for both tornadoes and hurricanes using Video Restoration Transformer (VRT) \cite{liang2022vrt} for super-resolution. It enhances each image from 1024 x 1024 pixels to 4096 x 4096 pixels, while maintaining the relative building locations and proportions. The YOLOv11 model, trained on customized xView datasets (building category only), is then used for building detection on the upscaled pre-disaster images. Leveraging the bounding boxes information generated from the YOLOv11 prediction, we cropped both pre- and post-disaster images around the predicted building areas with 30\% additional padding to prevent the truncation issue. This is a crucial step to provide only the precise information about the targeted buildings to the VLM for damage analysis. It reduced the information volume that the VLM processes, enabling resources to be allocated more effectively to the important areas. The reason for selecting the YOLOv11 model for our experiment is that YOLOv11 has been widely used in building detection and segmentation tasks and has outperformed other counterpart models \cite{el2025mask, zhao2025mar}. Each predicted building's cropped section from pre-and post-images is supplied to the VLM model, along with a curated prompt, to classify the severity of building damage according to the xBD dataset standards, and generate some useful information for the rescue teams, like the nature of hazards, characteristics of damaged buildings, and essential safety recommendations for the rescue workers. The generated VLM analysis helps natural disaster recovery workers to prioritize securing dangerous buildings at the highest category level to prevent further damage. The workflow of the proposed multi-stage AI framework for structural damage assessment is detailed in Figure \ref{fig:flow_diagram}. The xBD dataset categorizes building damage into four levels: (1) No/Slight Damage; (2) Moderate Damage; (3) Severe Damage; and (4) Totally Destroyed.

\section{Experimental Details}
%\ahmed{Overall remarks: We have a strong background of the thing that we are doing. However, the results section we are significantly weak to show any concrete results. Why we used SR and how it improves any of the results is also missing.}

\subsection{Experimental Configuration}
%\bijay{A paragraph of the machine setup you used for the experiment (GPU, Python version, CUDA version, framework used, any hyperparameter threshold used like used w = 2.5 for CLIPScore.}
For this experiment, we utilized a system with a Ryzen 9 9900X3D CPU, Nvidia 5070 GPU (12Gb GDDR7), and 128Gb DDR5 RAM at 4800MHz. We used Python version 3.14, and employed VRT with the videosr\_bi\_REDS model for video super resolution, targeting bicubic degradation reversal while training on the Resolution Enhancement in New Domains with Second-order Image Derivatives dataset. A CLIPScore weighted value of 2.5 was applied to uniformly stretch results from 0 to 1.

\begin{figure}[!h]
    \centering
    \includegraphics[width=\linewidth]{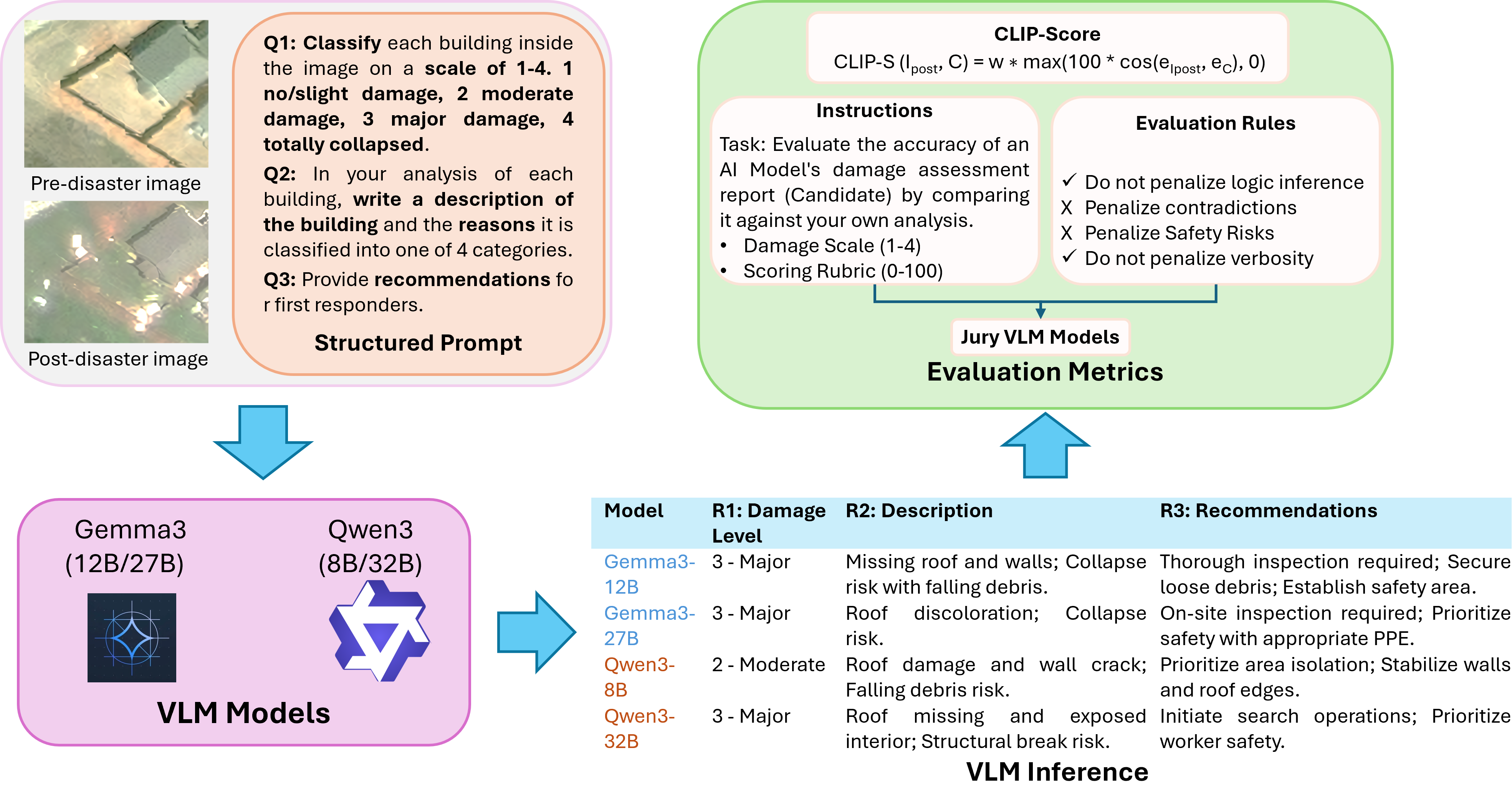}
    \caption{Overview of the Multi-VLM framework for disaster damage assessment. The framework provides pre-and post-disaster images with prompt as input to the various VLM models (Gemma3 and Qwen3). The generated responses from each model are evaluated using CLIPScore and VLM as a Jury metrics to assess the reasoning quality.}
    \label{fig:vlm_analysis}
\end{figure}

\subsection{Experimental Pipeline for Damage Assessment}

%\ahmed{Provide the experimental setup}

%\bijay{You can summarize the experimentation you did for VLM evaluation (discuss the prompt used, response format, model used, detailed evaluation method like averaging tokens for CLIPScore.)}

For damage assessment, four VLM models are utilized, which include two variants of Gemma3 (12 billion and 27 billion parameters) \cite{gemma2025gemma3} and two variants of Qwen3-vl (8 billion and 32 billion parameters) \cite{bai2025qwen3vl}. Here, Gemma3 variants represent the non-thinking models, while Qwen3-vl variants represent the models with thinking capabilities. Each VLM model is provided with the curated prompt, along with cropped pre- and post-disaster buildings to categorize the structural damage to the post-disaster buildings and the reason for the assigned damage level, based on the four categories adopted by the xBD dataset. The damage analysis response generated by each VLM is then evaluated using two metrics: CLIPScore and VLM-As-A-Jury to assess the accuracy and reasoning of classification. The CLIPScore metric has a limitation of its restriction to analyze only 77 tokens per caption, while our VLM generates comparatively large captions. To address this limitation, we calculate the CLIPScore for each 77-token segment of our VLM output and then average the scores, also noting the minimum and maximum scores of the individual segments. To evaluate our model comprehensively, we employ a dual-metric approach that includes a reference-free caption comparison and a visual comparison of VLMs. We first utilize a CLIPScore to evaluate our VLM output by examining images before and after a disaster, and then the information is used to measure the semantic alignment of a generated image caption regarding the damaged buildings~\cite{hessel2022clipscorereferencefreeevaluationmetric}.

To further validate the accuracy of the damage assessment, we employed the strategy called VLM-As-A-Jury to reduce bias often associated with the single model method. by using 4 different models we are able to evaluate outputs from candidate models, grading their performance based on a damage scale and a scoring rubric that varies communication effectiveness. Each model grades outputs on a scale from 0 to 100, producing a comprehensive .json format report detailing scores, classification accuracy, and a short reasoning for the specific damage score. The overall rankings for each candidate model are determined by averaging the scores provided by the jury members, thereby ensuring a nuanced evaluation across different model families. Figure  \ref{fig:vlm_analysis} illustrates the overall pipeline of the multi-VLM disaster damage assessment used in our research work.

\subsection{Evaluation Metrics}

\textbf{CLIPScore.} CLIPScore is a reference-free evaluation metric designed for evaluating image-text alignment, especially for assessing image captions without requiring human-generated reference captions \cite{hessel2022clipscorereferencefreeevaluationmetric}. A pretrained CLIP model (VIT-B/32) encodes both images and candidate captions into embedding vectors. The CLIPScore measures similarity, typically using cosine similarity, between these two vectors. A higher similarity indicates that the caption is more consistent with the image content.

Given a pair of images $I_{pre}$ (a pre-disaster image) and $I_{post}$ (a post-disaster image), the cosine similarity is calculated using the following equation:
\begin{equation}
    \text{CLIPScore}(I_{post}, C) = w \cdot \max(100 \cdot \cos(e_{I_{post}}, e_C), 0)
\end{equation}

where $e_{I_{post}}$ and $e_C$ represent the image and text embeddings, respectively, and w is a constant scaling factor used. We compute the score primarily against $I_{post}$ to ensure the model accurately describes the resulting damage. \\\\
\begin{figure} [!b]
    \centering
    \includegraphics[width=1.0\linewidth]{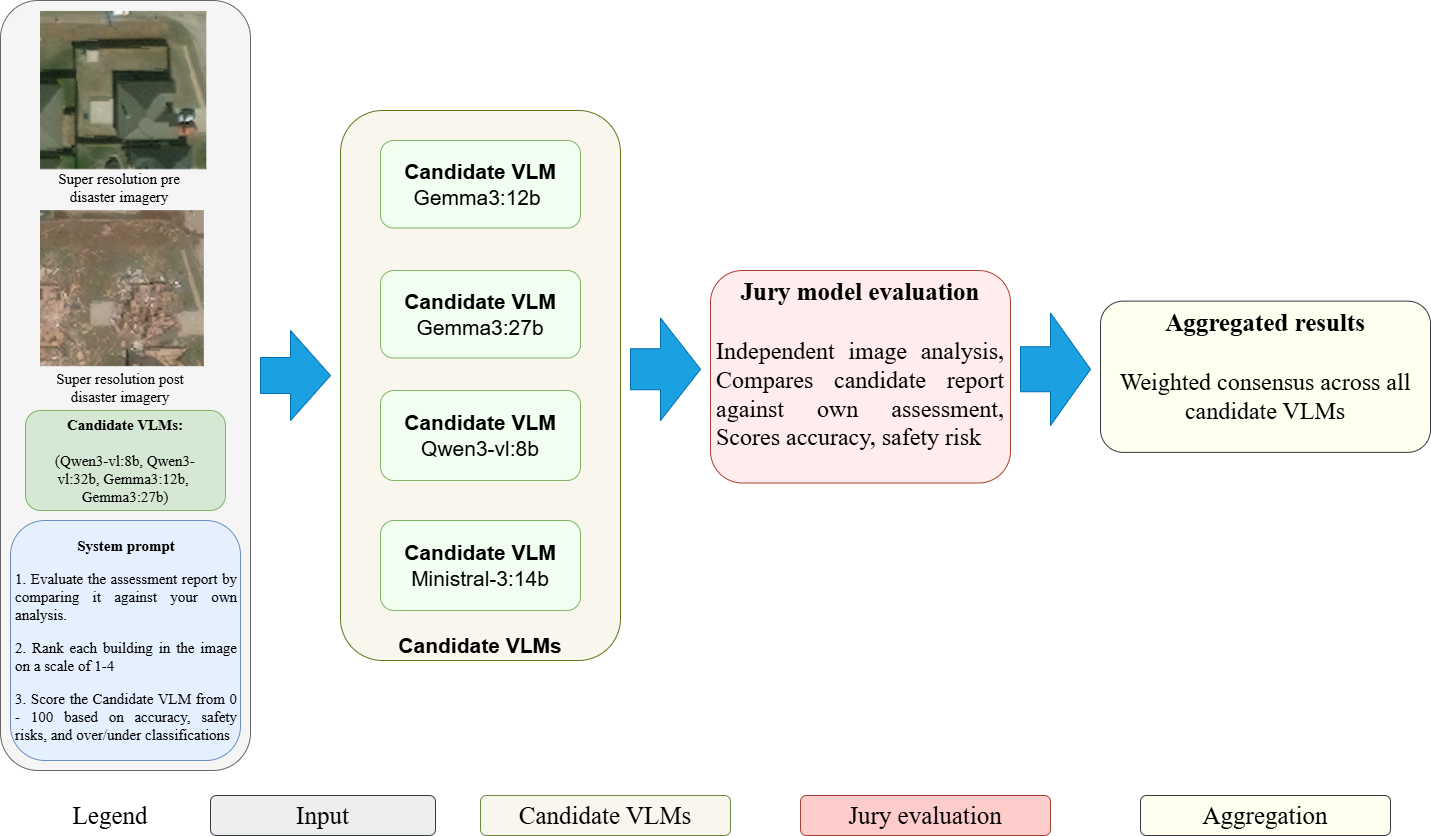}
    \caption {Working mechanism of the VLM-As-A-Jury metric.}
    \label{fig:workflow_jury}
\end{figure}
\textbf{VLM as a Jury.} VLM as a Jury is an evaluation metric that utilizes a robust VLM model to evaluate the outputs from the other candidate VLM models. Rather than comparing the generated response against a potentially flawed ``ground truth", it uses an input image, a user prompt, and candidate model responses to assess the VLM response based on various factors such as correctness, completeness, relevance, clarity, and hallucination risk. It is widely used for tasks like damage assessment and visual QA, where multiple valid responses exist. Figure \ref{fig:workflow_jury} illustrates the high-level working mechanism of the VLM as a Jury metric. For our experimental evaluation, four jury member models are used, namely, Gemma3:12b, Gemma3:27b, qwen3-vl:8b, and Ministral-3:14b. The jury models comprise a variation of a VLM assessing its own score, another model from the same family, a model from a different family that performed initial ranking, and a model not previously tested on the same dataset.

Each of the four jury members evaluates the outputs from original candidate models, providing a score from 0 to 100. The jury member model acts as a Senior Structural Engineer and Disaster Response Evaluator, scoring candidate VLMs based on a defined damage scale and scoring rubric. Scores range from 90-100 (excellent), 75-89 (good), 50-74 (weak), to below 49 (critical failure). The models provide results in a .json file, detailing the score, classification accuracy, reasoning for the score, the jury model, and the candidate model. The average scores from each jury member determine the overall rankings of the candidate models.

%\cat{The four jury member models include Gemma3:12b, Gemma3:27b, qwen3-vl:8b, and Ministral-3:14b. We use these in particular to see the judgments when the VLM is judging it's own score, a VLM in the same family of models, a VLM in a different family of models but also did the initial ranking, and a VLM which we have not tested with this dataset.}
%\ahmed{Who are the member models of the jury board? They are not mentioned in the paragraph.}

\section{Results}

\subsection{CLIPScore Evaluation on Full Images and Cropped Buildings}

\textbf{Full Image Evaluation.} Table \ref{tab:clipscore_full} assesses four VLM models for building damage assessment on two xBD disaster subsets, Moore Tornado, and Matthew Hurricane, using CLIPScore as the evaluation metric. The qwen3-vl:32b model leads with the highest average CLIPScores of $63.34$ and $62.42$ for the tornado and hurricane datasets, respectively, with a highest maximum CLIPScore of $81.04$ on the Matthew Hurricane dataset. In contrast, the VLCE baselines (LLaVA and QwenVL) lag behind at scores of $55.34$ and $60.60$, indicating the proposed framework is effective for specialized tasks like disaster scene understanding.

The finding shows that Gemma3:27b underperforms qwen3-vl:8b despite more parameters, scoring $60.02$ compared to $62.87$ on the Moore Tornado subset and $58.18$ compared to $62.17$ on the Matthew Hurricane subset. The results indicate that model size is not the sole determinant of performance, architecture also plays a crucial role. Additionally, Gemma3:27b and Gemma3:12b have nearly identical scores of $60.02$, reflecting comparatively diminishing performance compared to the Qwen3 variants for this type of task. 

\textbf{Cropped Buildings Evaluation.} For rigorous evaluation, the four VLM models are assessed on the cropped buildings subset, which is created from the bounding boxes information generated from the YOLOv11 buildings prediction. The result patterns remain the same as qwen3-vl:32b model outperforms the other counterparts on both disaster scenarios. According to the results highlighted in Table \ref{tab:clipscore_cropped}, the qwen3-vl:32b model achieves a CLIPScore of $59.66$ on the Moore Tornado subset and $58.11$ on the Matthew Hurricane subset. The results demonstrate poorer performance on the cropped building subset than on the full image dataset. 

All models except Gemma3:27b model perform better than the baseline VLCE (LLaVA) model, while lagging behind the VLCE (QwenVL) performance. One of the most notable anomalies witnessed in Table \ref{tab:clipscore_cropped} is that Gemma3:12b model yields the highest maximum CLIPScore of $70.24$ in the Hurricane dataset, which indicates that Gemma3:12b may generate well-aligned reasoning for certain patches, though the overall inconsistency is evidenced by a low average score of $56.56$ and a minimum of $42.14$. The overall evaluation highlights that all four VLM models face challenges in the cropped buildings subset due to fine-grained patch image inputs. The performance degradation is likely caused by the low-quality patches generated when cropping specific building areas from the full image.

\subsection{VLM-As-a-Jury Evaluation on Full Images and Cropped Buildings}

\textbf{Full Image Evaluation.} The performance comparison between Qwen3-VL and Gemma3 models reveals a significant gap. Qwen3-vl:8b scores $93.93$ on the Moore Tornado, slightly higher than Qwen3-vl:32b at $93.33$. In contrast, the Qwen3-vl:32b model excels in Hurricane Matthew with 90.22 over 88.66 for the 8b version. Both Qwen models demonstrate strong effectiveness, achieving on-par results. In contrast, Gemma3 models lag, achieving scores of $79.50$ and $79.61$ for Gemma3:27b, and lower for Gemma3:12b at $75.81$ and $71.33$, highlighting their inferior capability.

Analyzing the performance across different disaster subsets as presented in Table \ref{tab:combined_jury_results}, both the Qwen and Gemma models perform poorly in the Matthew Hurricane dataset compared to the Moore Tornado subset. This suggests that Hurricane Matthew presents a more challenging environment, like tiny objects and varying altitudes. Therefore, the disaster type and visual characteristics play a crucial role for assessing the performance of different models in disaster assessment. 

\textbf{Cropped Buildings Evaluation.} The results from VLM-as-a-Jury demonstrate a significant distinction between the Qwen3-VL and Gemma3 families even in the cropped building dataset. As specified in Table \ref{tab:cropped_images_jury_results}, Qwen3-vl:32b scores highest on both Moore ($88.60$) and Matthew ($87.64$), closely followed by Qwen3-vl:8b at $88.58$ and $87.23$, respectively. This indicates that Qwen models excel not only in image-text similarity but also in response quality, reasoning, and damage interpretation. Conversely, Gemma3 models lag, particularly on the Matthew Hurricane subset, with Gemma3:12b scoring only $67.93$. This reveals that Qwen models provide more reliable and consistent damage assessments in disaster scenarios, while Gemma models are less effective, especially in complex situations.

Compared to the evaluation on the full image, the performance of all four VLM models on the cropped buildings shows a consistent pattern, supporting the reliability of the overall findings. The Qwen3-vl models continue to excel, with Qwen3-vl:32b outperforming Qwen3-vl:8b in both the Moore Tornado and Hurricane Matthew subsets, although the difference is minimal. Both evaluations indicate that the Qwen3-VL family is the strongest, with the 32B version being particularly balanced. Additionally, Hurricane Matthew presents a greater challenge, causing lower scores for all models, most notably for Gemma3 variants. This alignment between both evaluations outlines that Qwen3-vl:32b is the most reliable model, while Qwen3-vl:8b offers nearly equivalent damage assessment quality.

\begin{table*}[!h]
    \centering
    \small
    \caption{CLIPScore comparison of the proposed framework using four different VLM models (baseline) on the upscaled xBD disaster subset (full images). The table reports the CLIPScore (\%) obtained by two VLCE variants: LLaVA-based and Qwen-VL–based as baselines.}
    \label{tab:clipscore_full}
    \begin{tabular}{l c c c c}
    \hline
    \textbf{Disaster type} & \textbf{VLM model} & \textbf{Avg. CLIPScore} & \textbf{Max. CLIPScore} & \textbf{Min. CLIPScore} \\ \hline
    \multirow{2}{*}{\textbf{xBD}} & VLCE (LLaVA-baseline) \cite{rahman2025vlce} & 55.34 & - & - \\
     & VLCE (QwenVL-baseline) \cite{rahman2025vlce} & 60.60 & - & - \\
    \hline
    \multirow{4}{*}{\textbf{Moore Tornado}}& Qwen3-vl:32b & \textbf{63.34} & \textbf{72.60} & \textbf{54.83} \\
        & Qwen3-vl:8b  & 62.87 & 70.42 & 51.40 \\
        & Gemma3:27b   & 60.02 & 70.69 & 50.23 \\
        & Gemma3:12b   & 60.02 & 68.55 & 51.80 \\ \hline
    \multirow{4}{*}{\textbf{Matthew Hurricane}} & Qwen3-vl:32b & \textbf{62.42} & \textbf{81.04} & 50.18 \\
        & Qwen3-vl:8b  & 62.17 & 77.56 &  \textbf{51.60} \\
        & Gemma3:27b   & 58.18 & 67.72 & 47.19 \\
        & Gemma3:12b   & 57.06 & 67.96 & 44.82 \\ \hline
\end{tabular}
\end{table*}

\begin{table*}[!ht]
    \centering
    \small
    \caption{CLIPScore comparison of the proposed framework using four different VLM models (baseline) on the upscaled xBD disaster subset (cropped building images).}
    \label{tab:clipscore_cropped}
    \begin{tabular}{l c c c c}
    \hline
    \textbf{Disaster type} & \textbf{VLM model} & \textbf{Avg. CLIPScore} & \textbf{Max. CLIPScore} & \textbf{Min. CLIPScore} \\  \hline
    \multirow{4}{*}{\textbf{Moore Tornado}} & Qwen3-vl:32b & \textbf{59.66} & \textbf{68.31} & 48.41 \\
    & Qwen3-vl:8b  & 58.79 & 66.50 & \textbf{50.30} \\
    & Gemma3:27b   & 56.70 & 63.40 & 48.10 \\
    & Gemma3:12b   & 56.58 & 67.78 & 47.79 \\ \hline
    \multirow{4}{*}{\textbf{Hurricane Matthew}} & Qwen3-vl:32b & \textbf{58.11} & 67.30 & \textbf{44.45} \\
    & Qwen3-vl:8b  & 58.08 & 66.38 & 42.78 \\
    & Gemma3:27b   & 55.03 & 62.44 & 42.94 \\
    & Gemma3:12b   & 56.56 & \textbf{70.24} & 42.14 \\ \hline
    \end{tabular}
\end{table*}

\begin{table}[!ht]
    \centering
    \caption{VLM-As-A-Jury evaluation comparison of the proposed method using four different VLM models (baseline) on the upscaled xBD disaster subsets (full image).}
    \label{tab:combined_jury_results}
    % This forces the table to fit the width of the text column
    \resizebox{\columnwidth}{!}{%
    \begin{tabular}{l c c c}
        \toprule
        \textbf{Candidate Model} & \textbf{Moore Tornado} & \textbf{Hurricane Matthew} \\
        \midrule
        Qwen3-vl:32b & 93.33 & \textbf{90.22} \\
        Qwen3-vl:8b  & \textbf{93.93} & 88.66  \\
        Gemma3:27b   & 79.50 & 79.61  \\
        Gemma3:12b   & 75.81 & 71.33 \\
        \bottomrule
    \end{tabular}%
    }
\end{table}

\begin{table}[!ht]
    \centering
    \small
    \caption{VLM-As-A-Jury evaluation comparison of the proposed method on upscaled xBD disaster subsets (cropped buildings).}
    \label{tab:cropped_images_jury_results}
    \begin{tabular}{l c c}
        \hline
        \textbf{Candidate Model} & \textbf{Moore Tornado} & \textbf{Matthew Hurricane} \\ \hline
        Qwen3-vl:32b & \textbf{88.60} & \textbf{87.64} \\
        Qwen3-vl:8b  & 88.58          & 87.23 \\
        Gemma3:27b   & 76.03          & 73.42 \\
        Gemma3:12b   & 75.65          & 67.93 \\ \hline
    \end{tabular}
\end{table}

\subsection{Ablation Evaluation without Super-resolution}

In order to determine the effect of super-resolution on the damage assessment tasks, we evaluated our framework without applying VRT super-resolution on the input images. Due to the computational limitations, we evaluate only the small variants of each VLM family: Qwen3-vl:8b and Gemma3:12b using CLIPScore metrics as depicted in Table \ref{tab:clip_scores_comparison_non_sr}. Even before applying super-resolution, Qwen3-vl:8b outperforms Gemma3:12b on both disaster subsets, scoring 60.52 and 60.04 on Moore Tornado and Hurricane Matthew, respectively, compared to Gemma3:12b's scores of 58.92 and 57.34.

Comparing the results in Table \ref{tab:clipscore_full}, it is proven that applying the super-resolution shows better performance for both disaster subsets. Qwen3\-vl:8b improved from 60.52 to 62.87 on Moore and from 60.04 to 62.17 on Matthew, marking an improvement of $+2.35$ on Moore and $+2.13$ on Matthew. Gemma3:12b also increased to 60.02 on Moore but only slightly decreased to 57.06 on Matthew, resulting in a $+1.10$ gain on Moore but a decline of $-0.28$ on Matthew. The results show that the super-resolution technique benefits the Qwen3 model more consistently than the Gemma3 model.

%In order to determine the effect that super-resolution images have on the natural damage assessment, we evaluate our methodology without the super-resolution step using VRT. We evaluate this using both the CLIPScore and VLM as a jury to ensure proper testing and accuracy.We can see that when compared to proper usage of our methodology the super resolution of images does enhance and increase the overall accuracy of the damage assessment when using both the CLIPScoring as well as the VLM as a jury strategy. Due to the high risk situations post natural disasters, this improvement can bring significant safety improvements for both natural disaster recovery personnel as well as civilians affected by the natural disaster itself.

\begin{table}[!ht]
\centering
\caption{CLIPScore comparison between Moore Tornado and Hurricane Matthew before super-resolution (full images).}
\label{tab:clip_scores_comparison_non_sr}
\begin{tabular}{l c c}
\toprule
\textbf{VLM Model} & \textbf{Moore CLIPScore} & \textbf{Matthew CLIPScore} \\ \midrule
Qwen3-vl:8b     & \textbf{60.52}                             & \textbf{60.04}                       \\ 
Gemma3:12b      & 58.92                             & 57.34                       \\ \bottomrule
\end{tabular}
\end{table}

%\begin{table}[!ht]
%\centering
%\caption{CLIPScore and VLM-as-a-Jury analysis on the Moore Tornado disaster subset before super-resolution.}
%\label{tab:clip_scores and VLM jury for non-sr images}
%\begin{tabular}{@{}lcc@{}}
%\toprule
%\textbf{Model} & \textbf{CLIP Score} &\textbf{VLM as a Jury} \\ \midrule
%Qwen3-vl:8b & 60.52 & 91.78 \\
%Gemma3:12b & 58.92 & 73.26\\ 
% \bottomrule
%\end{tabular}
%\end{table}
%TODO ADD THE VLM AS JURY TO THIS SPECIFIC TABLE

%\cat{To determine the optimal configuration for image enhancement, we evaluated X super-resolution (SR) algorithms against a non-enhanced control baseline. Performance was measured using the CLIPScore metric, which provides an objective, automated benchmark for image-text alignment, thereby eliminating the subjectivity and variance inherent in human-generated captions.}
%TABLE OF ALGORITHMS AND THEIR CLIPScore

\subsection{Building Detection Performance (VLM and Ground Truth)}

To evaluate the accuracy of the VLMs against the ground truth, we established two analytical buckets leveraging the Moore Tornado subset. The first included categories 1 and 2, which represent minimal damage posing little to no threat to civilians or first responders after a natural disaster. The second bucket comprised categories 3 and 4, where significant risks exist due to highly damaged or completely collapsed buildings. It addresses a problem highlighted in the original xBD paper, where slight changes in categories lead to increased misclassification due to data imbalance in the natural disaster dataset. Most buildings are classified as ``no damage", and even minor label differences can result in lower classification results. We analyze the precision, recall, and F1-score of all four VLMs tested in our framework. Both the Qwen3 and Gemma3 models achieve decent accuracy of approximately 85\% - 87\%. The results presented in Table \ref{tab:building_performance} indicate that Qwen3-vl:32b is the top-performing model across all classification metrics, achieving the highest accuracy (87.1\%), precision (0.8198), recall (0.9342), and F1-score (0.8733). Qwen3-vl:8b closely follows with 86.8\% accuracy and an F1-score of 0.8652. In comparison, the Gemma3 models perform slightly worse compared to the Qwen3 variants, though the difference is smaller.

\begin{table}[!ht]
    \centering
    \small
    \caption{Performance evaluation of VLMs on building detection against ground truth.}
    \label{tab:building_performance}
    \begin{tabular}{l c c c c}
    \hline
    \textbf{VLM model} & \textbf{Accuracy(\%)} & \textbf{Precision} & \textbf{Recall} & \textbf{F1-Score} \\ \hline
    Qwen3-vl:32b & \textbf{87.1\%} & \textbf{0.8198} & \textbf{0.9342} & \textbf{0.8733} \\
    Qwen3-vl:8b  & 86.8\%          & 0.8074          & 0.9321          & 0.8652          \\
    Gemma3:27b   & 85.4\%          & 0.7931          & 0.9153          & 0.8498          \\
    Gemma3:12b   & 85.1\%          & 0.7874          & 0.9079          & 0.8434          \\ \hline
\end{tabular}
\end{table}

\subsection{Word Count Analysis of VLM-Generated Damage Descriptions}

The word clouds reveal a clear progression in language across the four damage categories, reflecting increasing severity in structural conditions. In Category 1, terms such as “quality” and “blurry” indicate uncertainty caused by limited visual evidence when damage is minimal. In Category 2, words like “moderate” and “roof” highlight localized structural damage, particularly to roof components.

\begin{figure}[!t]
    \centering
\includegraphics[width=1.0\linewidth]{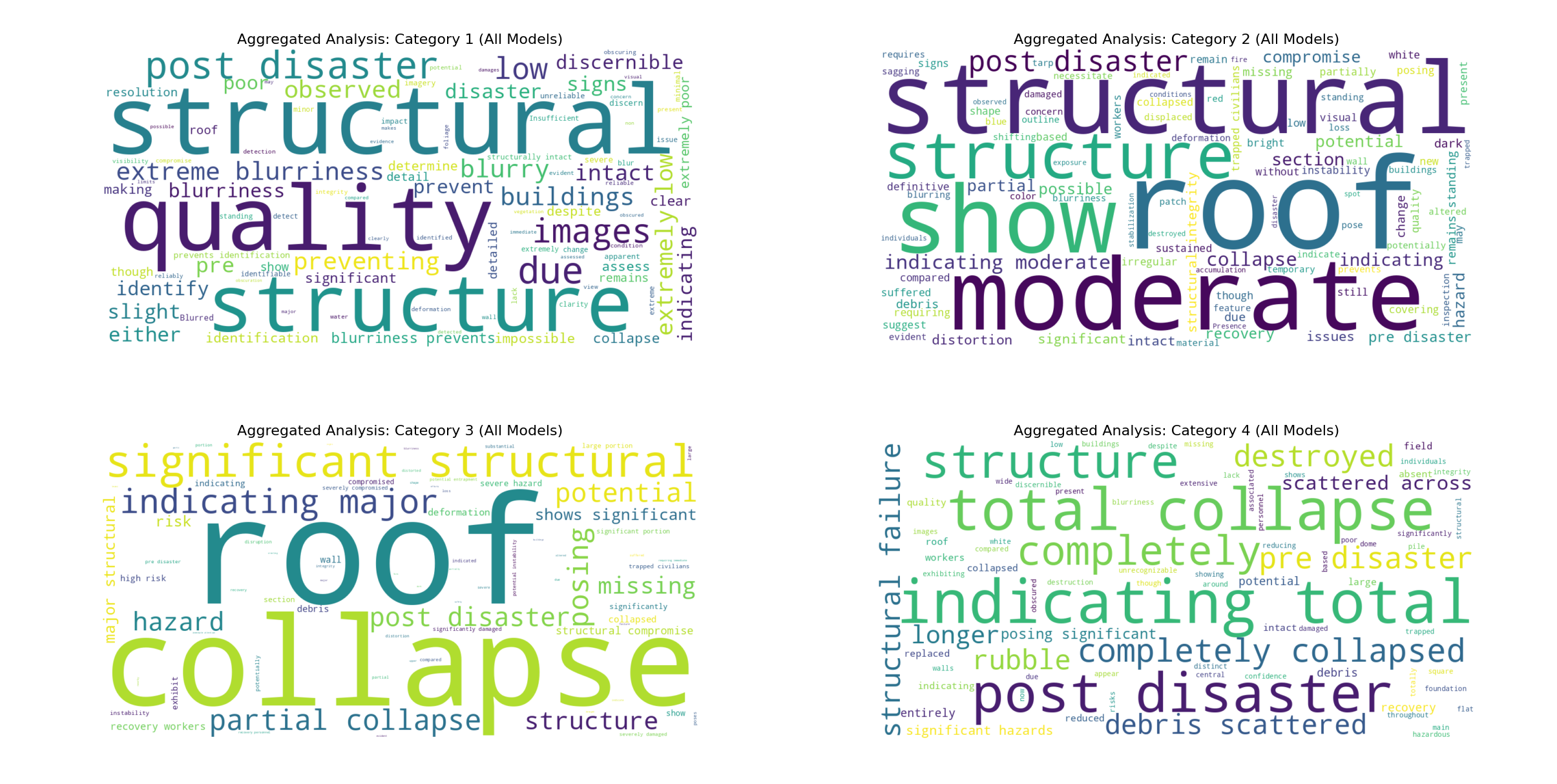}
    \caption{Aggregated word clouds of VLM-generated damage descriptions across the four damage categories.}
    \label{fig:word_cloud}
\end{figure}

In Categories 3 and 4, the language shifts toward severe damage. Category 3 includes terms such as “collapse” and “significant structural,” indicating major structural failure, while Category 4 features phrases like “total collapse” and “structural failure,” reflecting catastrophic destruction. This progression demonstrates that model-generated descriptions align with the expected hierarchy of disaster damage severity.
\section{Conclusion and Future Work}

We propose a hybrid framework that integrates AI-based super-resolution, YOLOv11 object detection, and VLMs for automated post-disaster damage assessment. The framework enhances satellite imagery from 1024×1024 to 4096×4096, addressing the limitations of low-resolution remote sensing data. The enhanced images enable more precise building localization using YOLOv11 and allow VLMs to perform more reliable damage analysis by reducing background noise and irrelevant contextual information. Experimental evaluation of open-weight VLMs using CLIPScore and a VLM-as-a-Jury strategy indicates that model architecture plays a more significant role in performance than parameter count. The Qwen3 family consistently outperformed the Gemma models (Gemma3:27B and Gemma3:12B), with Qwen3-VL (32B) achieving the best overall performance across both datasets, except for the minimum CLIPScore in the Hurricane Matthew dataset, where the Qwen3-VL 8B variant performed best. Furthermore, the multi-model jury approach effectively mitigates individual model bias and improves evaluation reliability in the absence of ground-truth captions.

Despite these promising results, several challenges remain. The current framework relies on computationally intensive components, including super-resolution algorithms and large-scale VLMs, which require substantial computational resources. Future work will focus on improving efficiency through lightweight models and optimized architectures. Additionally, the present study evaluates only two disaster types; extending the framework to other events such as earthquakes, floods, and wildfires remains an important direction. Finally, while the system provides recommendations for first responders based on detected damage, VLM performance can degrade in complex scenarios with poor image quality, highlighting the need for more robust reasoning mechanisms in future research.

{
    \small
    \bibliographystyle{ieeenat_fullname}
    \bibliography{main}
}

\end{document}